\documentclass[seceq,dvipdfmx]{article}

\usepackage{arxiv}

\usepackage[utf8]{inputenc} 
\usepackage[T1]{fontenc}    
\usepackage{hyperref}       
\usepackage{url}            
\usepackage{booktabs}       
\usepackage{amsfonts}       
\usepackage{nicefrac}       
\usepackage{microtype}      
\usepackage{lipsum}
\usepackage{graphicx}
\usepackage{multirow}
\usepackage{tabularx}
\usepackage{amsmath}
\graphicspath{ {./images/} }
\usepackage[numbers]{natbib}

\usepackage{algorithm}
\usepackage{algorithmic}
\usepackage{color}
\usepackage{booktabs}
\usepackage{enumitem}

\usepackage{url,hyperref,lineno,microtype,subcaption}
\usepackage{tikz}
\usetikzlibrary{arrows}
\usepackage {booktabs}
\usepackage{tikz}
\usetikzlibrary{arrows,calc,chains}
\makeatletter
\DeclareRobustCommand{\rvdots}{%
  \vbox{
    \baselineskip4\p@\lineskiplimit\z@
    \kern-\p@
    \hbox{.}\hbox{.}\hbox{.}
  }}
\makeatother

\setlist[enumerate]{topsep=3pt,parsep=0pt,partopsep=0pt,itemsep=0pt,leftmargin=2zw,labelsep=*}

\title{Constitutive Components for Human-Like Autonomous Artificial Intelligence}

\author{
Kazunori D Yamada \\
yamada@tohoku.ac.jp\\
Tohoku University\\
Sendai, Japan\\
}

\begin{document}
\maketitle
\begin{abstract}
This study is the first to clearly identify the functions required to construct artificial entities capable of behaving autonomously like humans, and organizes them into a three-layer functional hierarchy. Specifically, it defines three levels: Core Functions, which enable interaction with the external world; the Integrative Evaluation Function, which selects actions based on perception and memory; and the Self Modification Function, which dynamically reconfigures behavioral principles and internal components. Based on this structure, the study proposes a stepwise model of autonomy comprising reactive, weak autonomous, and strong autonomous levels, and discusses its underlying design principles and developmental aspects. It also explores the relationship between these functions and existing artificial intelligence design methods, addressing their potential as a foundation for general intelligence and considering future applications and ethical implications. By offering a theoretical framework that is independent of specific technical methods, this work contributes to a deeper understanding of autonomy and provides a foundation for designing future artificial entities with strong autonomy.
\end{abstract}

\section{Introduction}
\subsection{Background}
Constructing autonomous artificial entities is one of the fundamental goals of artificial intelligence research. While the performance of artificial intelligence systems specialized for specific tasks has improved dramatically, most of them operate based on externally defined goals. Their ability to choose or update actions based on their own internal states or past experiences remains limited \cite{bubeck2023, lake2017}.

Humans, by contrast, perceive their environment, accumulate experiences, and behave according to the situation. Their actions are not merely reactive but are chosen based on meaningful judgment, and they possess the flexibility to change their behavioral policies or values when needed \cite{franklin1997, totschnig2020}. Artificially constructing such human-level entities remains an unsolved challenge in current artificial intelligence research. At the same time, it represents an important step toward creating entities that can be treated as others in social interaction.

\subsection{Behavior and Autonomy}
The autonomy of artificial entities is a key concept that significantly affects both the quality of their behavior and their acceptance in society. There are two levels of autonomy: weak autonomy and strong autonomy \cite{totschnig2020}.

Weak autonomy is the ability to choose and carry out actions based on one's own judgment under externally defined goals. Many current artificial intelligence systems are designed to perform specific tasks and are capable of selecting optimal means to achieve given objectives. However, they are not able to set their own goals.

In contrast, strong autonomy is the ability to set one's own goals and values, and to select and carry out actions based on them. Entities with this level of autonomy can determine their behavioral policies internally and flexibly revise them when necessary.

As such, the level of autonomy fundamentally changes their behavior and their acceptance in society. This paper builds on this distinction and discusses the functional components essential for entities to behave autonomously.

\subsection{Research Objectives}
The goal of this study is to explicitly identify the necessary and independent functions required to construct artificial entities that behave autonomously like humans, and to organize them into a three-level functional hierarchy. This paper defines the core components that support autonomy as the following three functional groups:
\begin{itemize}
  \item Core Functions,
  \item Integrative Evaluation Function,
  \item Self Modification Function.
\end{itemize}
These represent, respectively: functions that enable interaction with the external world; functions that select actions based on perception and memory; and functions that reconfigure not only evaluation policies and behavioral principles, but the structure of all functions within the artificial entity itself.

Through this framework, the study aims to offer a constructivist and implementation-oriented perspective on the philosophical question of what autonomy is, while also providing clear guidance for future artificial intelligence design.

\section{Related Works}
\subsection{Engineering and Philosophical Definitions of Autonomy}
Autonomy is a multifaceted concept in artificial intelligence research, and its definition differs significantly between philosophical and engineering perspectives.

From an engineering perspective, autonomy is often understood in an operational and gradual manner \cite{franklin1997, nilsson1994}. For example, in the design of autonomous mobile robots, the ability to act in response to the environment without continuous external control is considered autonomy. This is referred to as weak autonomy, where the agent can independently choose means to achieve goals that are externally given \cite{totschnig2020}.

In contrast, from a philosophical perspective, autonomy is defined as the capacity of an agent to determine its own actions without any external interference, and at its core lie concepts such as free will and moral judgment \cite{buss2018, audi1991}. A well-known example is Kant, who defined autonomy as self-legislation, the idea that a moral agent must internally determine the principles of its own actions \cite{kant1785, denis2005}. Autonomy from this perspective is referred to as strong autonomy, which includes the ability to define one's own goals and values and to flexibly restructure one's standards of evaluation \cite{totschnig2020}.

Thus, autonomy should not be seen as a single function but rather as a continuous spectrum, ranging from the ability to act toward external goals to the ability to generate and revise goals themselves. When designing artificial entities, the level of autonomy being targeted determines the required functions and design principles.

\subsection{Consciousness, Value, and Ethics in Autonomy}
Building strong autonomy requires more than just behavioral flexibility; it necessitates an understanding of self-generated goals. In this context, issues related to consciousness, values, and ethics play a central role.

For example, in theories of consciousness such as Global Workspace Theory and Higher-Order Thought Theory, the integration of information and selective access are thought to produce higher-order representations of the self, which are linked to behavioral control \cite{baars1988, farrell2018}. Such cognitive architectures may serve as a foundation for strong autonomy, as they enable explicit representation and evaluation of one's own goals and values.

In addition, the design of reward functions and the evaluation of value in artificial intelligence are closely related to autonomy \cite{russel2010}. In traditional reinforcement learning, behavior is guided by a predefined value function. However, discussions in ethical artificial intelligence argue that the meaning and justification of behavior should be re-evaluated based on relationships with others and internal values \cite{abel2016}. To address this, approaches have been proposed that use inverse reinforcement learning to learn human values, or that reference explicit ethical principles as part of the decision process \cite{noothigattu2019}.

\subsection{Positioning of This Study}
This study, rather than focusing only on conceptual frameworks and ethical demands found in prior research on building artificial entities with human-level strong autonomy, clearly identifies the necessary functional components and organizes them into a hierarchical model.

First, it positions four fundamental functions: Perception, Memory, State Description, and Motor Execution, as the most essential capabilities required for entities that exhibit human or human-like behavior.

Next, it identifies the Integrative Evaluation Function as the core of human-likeness, distinguishing such entities from merely reactive machines. This function integrates perceptual and memory information to select meaningful actions.

Furthermore, it proposes that the Self Modification Function enables higher performance and flexible restructuring of goals and behavioral principles. Through this function, strong autonomy can be realized in a deeper sense, including ethical judgment and the reconstruction of values.

\section{Functional Architecture of Autonomous Artificial Intelligence}
\subsection{Ontology in the Second Person Perspective}
In discussions of autonomy in artificial intelligence, questions about the presence or absence of consciousness, thought, and emotion have frequently been raised \cite{brunette2009, yamada2022, martinez2005, reggia2020, searle1980}. However, from a second-person perspective, recognition as an other can be established even without assuming such internal states \cite{jaegher2010}. For example, the philosophical zombie argument suggests that consciousness cannot be verified from a third-person standpoint and may not be essential in social relationships \cite{chalmers2004}.

From this standpoint, whether an entity is considered autonomous depends on whether its behavior enables meaningful interaction with others. Consistent and adaptive behavior in response to the environment and context is what makes an entity recognizable as socially other.

Therefore, in constructing autonomous artificial entities, the key design requirement is not the presence of consciousness or intention, but rather the structure of behavior that enables social interaction. Based on this view, the following sections examine the minimal functional composition required for an artificial entity to behave like a human.

\subsection{Fundamental Functional Architecture of Autonomous Entities}
To construct autonomous artificial entities that behave like humans, it is not necessary to assume internal cognitive features such as reasoning or consciousness. Pascal once described humans as \textit{a thinking reed}, seeing thought as the essence of humanity. However, as discussed in the previous section, from a second-person perspective, whether or not an entity is thinking is not the point. What matters is the observable structure of behavior and the capacity for interaction with others.

From this standpoint, this study categorizes the minimal set of functions required for autonomous behavior into three levels and explicitly defines each function in a constructive manner.

At Level 1, we define Core Functions as consisting of only four basic capabilities necessary for interaction with the environment: Perception, Memory, State Description, and Motor Execution.

When a human is observed from a second-person perspective as if they were a machine, they appear to be entities that sense information from the environment, store that information, produce output in the form of language, and physically affect the world. Whether such an entity imagines, thinks, is conscious, or has emotions cannot be known from this perspective and does not need to be considered. These four functions are precisely the foundational capabilities that humans appear to exhibit when observed second-personally.

At Level 2, the Integrative Evaluation Function is added. This function integrates the perceptual and memory information to produce meaningful judgments.

At Level 3, the Self Modification Function is introduced. It allows the system to enhance its capabilities by modifying the Core Functions, and to enable the generation or revision of goals by modifying the Integrative Evaluation Function.

As shown in Table \ref{tab:functions}, these functions have distinct roles. When integrated in a stepwise manner, they enable the realization of more advanced forms of autonomy.

    \begin{table}[h]
    \centering
    \begin{tabularx}{\textwidth}{llX}
    \toprule
    \textbf{Functional Level} & \textbf{Function Name} & \textbf{Description} \\
    \midrule
    \multirow[t]{4}{*}{Level 1} 
    & Perception Function & Senses the state of the external world and the self \\
    & Memory Function & Enables storage and reuse of experiential information \\
    & State Description Function & Describes situations and memory content \\
    & Motor Execution Function & Physically affects the environment \\
    Level 2 & Integrative Evaluation Function & Evaluates and integrates perceptual and memory information to decide actions \\
    Level 3 & Self Modification Function & Dynamically restructures the architecture of the Core Functions and the Integrative Evaluation Function \\
    \bottomrule
    \end{tabularx}
    \vspace{0.6em}\\
    \caption{Constituent functions required for autonomous artificial entities}
    \label{tab:functions}
    \end{table}

\subsection{Core Functions}
\subsubsection{Perception Function}
The Perception Function is the function responsible for receiving input information in an autonomous artificial entity. It serves as sensing the state of the environment as well as the internal state of the entity itself. This function detects both physical and social stimuli from the external world and incorporates them in a processable internal format. It also includes the ability to sense internal information, which relates to mechanisms known as self-monitoring.

The Perception Function precedes all other functions, as it is necessary for acquiring the fundamental information required for action selection, state description, and higher-level evaluation.

\subsubsection{Memory Function}
The Memory Function enables an autonomous artificial entity to accumulate past experiences and retain them in a reusable form. It is the ability to preserve information over time and to retrieve previously acquired knowledge. This function not only stores information but also enables the retrieval of relevant memories when needed.

Various forms of memory can be assumed, including short-term storage, long-term knowledge accumulation, and structural distinctions such as episodic and semantic memory. However, this paper does not address such details. Instead, it focuses on the function of storing perceived information in a form that allows later reference and use.

The Memory Function is essential for linking past experience to the current situation, enabling the artificial entity to exhibit coherent behavior.

\subsubsection{State Description Function}
The State Description Function is the function that transforms perceptual and memory information into a representable format. Internally, this description is used to maintain a structured representation of the current situation. Externally, it is expressed as linguistic output, forming the basis for communication with others.

This function does not evaluate or select the input information. Rather, it serves as simply verbalizing the perceived facts or memory content. For example, describing the object in front as a red sphere or verbalizing a past event are typical operations of this function.

In this way, the State Description Function is one of the most fundamental output functions of an autonomous artificial entity. Before any organization or interpretation of the input information takes place, it performs the role of expressing the situation in a standardized manner that can be shared with others.

\subsubsection{Motor Execution Function}
The Motor Execution Function is the function responsible for translating the result of decision-making into physical output that acts upon the environment. This function serves as executing movements themselves.

Its essence lies in producing physical output, and it does not include mechanisms for selection or evaluation. For example, executing motor actions such as walking, grasping objects, or extending an arm all fall under this function. The evaluation of output appropriateness and the selection of actions are delegated to a higher-level function.

In this way, the Motor Execution Function serves as the final output mechanism through which an autonomous artificial entity exerts influence on its environment. It is a fundamental component that enables behavior as a physically embodied agent.

\subsection{Integrative Evaluation Function}
The Integrative Evaluation Function is the function responsible for processing perceptual and memory information in an integrated manner to select contextually appropriate actions. Without this function, behavior lacks coherence and purpose in relation to the environment and becomes nothing more than random output. In other words, the presence of this function is essential for generating meaningful behavior and for realizing human-level performance.

In the absence of this function, each of the Core Functions operates independently. As a result, recognition and action, as well as memory and behavior, become disconnected, and behavioral consistency is lost. Even if the external environment is perceived, if that information is not reflected in the output, it cannot be regarded as a response. It is merely noise. Similarly, if accumulated experiences in memory are not utilized in behavior, the entity cannot be considered capable of learning.

Moreover, flexible adaptation to changing environments or situations cannot be achieved without this function. If an entity's behavior remains fixed despite changes in context, it cannot interact effectively with the environment and fails to respond appropriately to change.

In addition, if behavioral choices lack meaningful justification, it becomes impossible for others to understand why a particular action was chosen. This severely undermines the entity's recognition as an autonomous other from a second-person perspective. In social contexts, to understand another agent means to recognize that its behavior is guided by reasons or intentions.

For these reasons, the Integrative Evaluation Function can be defined as a mediating mechanism that links perception and memory through evaluation, assigns meaning to actions according to context, and selects the most coherent behavior. The existence of this function transforms mere physical output into deliberate action and enables it to be recognized as autonomous behavior.

\subsection{Self Modification Function}
The Self Modification Function is the capability to dynamically restructure the internal architecture of each Core Function to improve their performance, and to reconfigure the structure of the Integrative Evaluation Function in order to change behavioral strategies or even the goals themselves. This function enables an autonomous artificial entity to restructure itself through learning and development, supporting long-term adaptation to its environment.

Without this function, behavior becomes fixed, making it difficult to strategically update actions in response to environmental changes or accumulated experience. For example, if an entity that has been acting based on a certain reward structure needs to adapt to new values or social norms, it requires the capability to reconstruct its evaluative criteria. Such reconstruction is not limited to parameter adjustments but involves a fundamental revision of the entire system of evaluation, decision-making, and output.

With the introduction of this function, an autonomous artificial entity can not only enhance its perceptual, memory, language, and motor capabilities, as well as improve its decision-making performance, but also continue to grow by reconstructing its evaluative criteria and goal structures. Therefore, the Self Modification Function is an indispensable component for realizing human-level strong autonomy.

\section{Hierarchical Structure and Behavioral Levels of Autonomy}
For an autonomous artificial entity to behave in a human-like manner, the functions described so far must not only operate individually, but also coordinate with one another in a hierarchical structure. This section explains how these functions interact to enable human-like behavior.

The Core Functions, namely Perception, Memory, State Description, and Motor Execution, constitute the minimal set required to support basic input and output with the environment, enabling reactive behavior. These functions receive sensory input, store information, generate linguistic output, and produce physical actions. However, when they operate independently, coherent behavior and learning become difficult.

With the addition of the Integrative Evaluation Function, perceptual and memory information can be evaluated and integrated, enabling context-sensitive decision making. At this stage, the entity's behavior aligns with environmental context and is recognized by others as meaningful.

Furthermore, the Self Modification Function enhances the performance of each Core Function and restructures the evaluation and decision-making mechanisms of the Integrative Evaluation Function. This enables flexible goal setting and the revision of value systems. Through this function, the entity transitions from weak autonomy to strong autonomy.

This hierarchical structure is not merely a sum of independent functions. Each function relies on the functions of the lower levels for its operation. For example, integrative evaluation is impossible without the foundational functions of perception, memory, and state description. Similarly, self-restructuring becomes feasible only when the Core Functions and the Integrative Evaluation Function are already in place.

Through this structure of interdependence, the consistency, adaptability, and meaningfulness of behavior are gradually acquired. Table \ref{tab:behaviors} summarizes how the stepwise expansion of this architecture corresponds to changes in behavioral quality. Each level represents not only an increase in the number of functions but also a change in the nature of behavior, such as how consistent, adaptive, or meaningful the actions become.

    \begin{table}[h]
    \centering
    \begin{tabularx}{\textwidth}{llX}
    \toprule
    \textbf{Behavior} & \textbf{Required Function Groups} & \textbf{Description} \\
    \midrule
    Reactive & Level 1 & A reactive entity that can respond to the external world through perception, memory, state description, and motor execution, but lacks coherence \\
    Weak Autonomous & Level 1, 2 & An autonomous entity (weak sense) capable of meaningful actions based on integrated evaluation according to the situation \\
    Strong Autonomous & Level 1, 2, 3 & An autonomous entity (strong sense) that continuously develops by reconstructing its own evaluation structure and goals \\
    \bottomrule
    \end{tabularx}
    \vspace{0.6em}\\
    \caption{Classification of structural levels in human-like autonomous artificial entities}
    \label{tab:behaviors}
    \end{table}

\section{Feasibility and Design Approaches}
\subsection{Reward Is Enough}
In recent years, a perspective that regards the reinforcement learning framework as the sole principle of intelligence has gained attention in the pursuit of artificial general intelligence. According to this view, all forms of intelligent behavior can ultimately be understood as the result of maximizing value. If the value function is appropriately designed, intelligence can form autonomously without the need for externally set goals or design interventions \cite{silver2021}. We support this perspective and focus on the possibility that a single principle of value maximization can give rise to diverse intellectual abilities.

This stance stands in contrast to conventional modular design approaches. It assumes that various intelligent abilities naturally emerge from the single principle of value maximization, and can be regarded as an extremely bold proposition. This view does not take the approach of explicitly constructing separate functional components, as presented in this study.

That said, even when growing a reinforcement learning agent based on the principle of value maximization, constructing an entity that behaves in a human-like manner requires a minimal set of functional components to be either initially built-in or acquired through learning. These functions are likely to be implicitly learned or utilized in the reinforcement learning process. They correspond to the Core Functions and the Integrative Evaluation Function proposed in this study.

On the other hand, there is a fundamental difference between this view and the concept of strong autonomy emphasized in this study. In reinforcement learning, the value function is typically given externally, whereas strong autonomy requires the agent to internally set and, when necessary, reconstruct its own goals and values corresponding to rewards. The current reinforcement learning framework appears to lack the Self Modification Function proposed in this study.

\subsection{Reinforcement Learning and Strong Autonomy}
Reinforcement learning is a framework in which agents learn action policies through interaction with the environment, based on the principle of value maximization. Agents in this framework are assumed to possess many fundamental cognitive abilities. For example, the ability to observe the state of the environment and perform actions corresponds to the Core Functions defined in this study: Perception, State Description, and Motor Execution. Furthermore, the process of selecting optimal actions based on past experience and current conditions can be understood as a policy, which fulfills the role of the Integrative Evaluation Function.

However, the capabilities that correspond to the Self Modification Function, such as developing or enhancing Core Functions and dynamically reconstructing evaluation criteria or behavioral strategies, are not sufficiently addressed by current reinforcement learning frameworks. For instance, if we wish to enable an artificial agent without language capabilities to utilize language, the agent must acquire a State Description Function. In general, reinforcement learning agents are not equipped with the ability to add new functions to themselves. Although techniques such as inverse reinforcement learning exist, which aim to infer reward structures, they are typically used to estimate human values externally. These methods differ fundamentally from the capacity of an agent to reassess its own objectives and value criteria and to update them autonomously \cite{ng2000, hadfieldmenell2016}. In this respect, reinforcement learning still lacks the capability to internally define and reconstruct objectives, which is a core requirement of strong autonomy.

Therefore, while reinforcement learning presents a promising framework for the implementation of strong autonomy, it does not inherently include all of the necessary components. In particular, the introduction of functions such as the Self Modification Function, which allow the agent to alter its values according to context and thereby realize meaningful behavior aligned with social contexts, will be an important area for future extension.

\subsection{Technical Prospects and Limitations of Implementation}
In this study, we defined the fundamental functions required to construct artificial entities with human-level strong autonomy and clarified their hierarchical structure. From the perspective of this constructive framework, reinforcement learning is a powerful approach that naturally realizes some of these functions.

However, reinforcement learning has limitations in supporting the Self Modification Function, which is essential for achieving strong autonomy. Addressing these limitations still requires technical and design-level interventions. In current reinforcement learning systems, the reward function is typically predefined by human designers. As a result, the goals and values that guide the agent's behavior are externally determined rather than autonomously constructed by the agent itself. This lacks the core condition for strong autonomy: the agent's ability to form and modify its own goals and values. Achieving such intrinsic goal construction may require an additional design layer that supports self-referential processing beyond standard reward-based learning. We argue that acquiring new values requires modifying the Integrative Evaluation Function via the Self Modification Function.

Recognizing these limitations, the constructive framework proposed in this study serves as a foundation for a stepwise implementation strategy that complements reinforcement learning. By supplementing reinforcement learning with the structurally defined functions identified here, it may become possible to design more robust and flexible autonomous artificial entities.

\section{Conclusion and Future Directions}
This study identifies the key functions necessary for building artificial entities capable of human-level autonomous behavior and organizes them hierarchically. The aim was to clarify the structural requirements for artificial systems to exhibit human-level capabilities. We adopted a second-person perspective, emphasizing the ability to behave in a human-like manner, rather than assuming the presence of consciousness or intent. By formalizing these functions instead of relying solely on philosophical, ethical, or design-based discussions, this work aimed to provide a foundational framework for constructive understanding and design.

The functions defined in this study, including Perception, Memory, State Description, and Motor Execution, are fundamental components required for an entity to behave in a human-like manner. Additionally, the Integrative Evaluation Function, which evaluates and integrates perceptual and memory information, plays a central role in establishing coherent interactions with the environment. Although not explicitly defined in many intelligent systems, it is likely an indispensable structure and may even correspond to what is commonly referred to as consciousness. Furthermore, the Self Modification Function, which enables the reconstruction of goals and decision-making frameworks, serves as a higher-level mechanism that supports the development and adaptation of the Core Functions and the Integrative Evaluation Function. It plays a critical role in the realization of strong autonomy.

This study focused on reinforcement learning as one possible implementation method because it offers promising potential to naturally derive diverse intelligent capabilities. However, reinforcement learning frameworks require the agent to be either equipped with or able to acquire the essential functions mentioned above. In the context of reinforcement learning, the Integrative Evaluation Function may be interpreted as being implemented in the form of a policy responsible for action selection. On the other hand, capabilities such as generating new Core Functions, modifying the Integrative Evaluation Function, altering one's own goals, or acquiring new values are not currently included in standard reinforcement learning architectures. Therefore, whether strong autonomy is pursued through reinforcement learning or through alternative approaches, a functional understanding of the components required to build autonomous artificial entities remains essential.

The framework presented in this study theoretically formalizes this functional structure and offers a theoretical foundation applicable to various technical approaches. Future work must aim to translate this structure into implemented systems that function within learning processes and social contexts. Together with ethical considerations and practical applicability, this framework is expected to serve as a theoretical foundation for designing and understanding artificial entities with human-level strong autonomy.

\section*{Acknowledgement}
This work was supported in part by the Top Global University Project from the Ministry of Education, Culture, Sports, Science, and Technology of Japan (MEXT).

\bibliographystyle{unsrt}
\bibliography{main}

\end{document}